%% file: main.tex
\pdfoutput=1
\documentclass[11pt]{article}

\usepackage{naacl2021}

\usepackage{times}
\usepackage{latexsym}

\usepackage{balance}
\usepackage[T1]{fontenc}
\usepackage[utf8]{inputenc}

\usepackage{microtype}

\usepackage{latexsym}

\usepackage{graphicx}
\usepackage{amssymb}
\usepackage{svrsymbols}

\usepackage{microtype}
\usepackage{multirow}

\usepackage{subcaption}

\usepackage{enumitem}

\newcommand{\stitle}[1]{\noindent{\textbf{#1}}}

\DeclareMathAlphabet{\mathpzc}{OT1}{pzc}{m}{it}

\usepackage[ruled,vlined]{algorithm2e}

\SetKwInput{KwInput}{Input}                
\SetKwInput{KwOutput}{Output}              
\SetKwInOut{Input}{Input}
\SetKwInOut{Output}{Output}
\SetKwInput{Parameter}{Parameter}
\SetKwInput{Parameters}{Parameters}
\SetKwInput{Algorithm}{Algorithm}
\SetKwInput{Remark}{Remark}
\SetKwInput{Variable}{Local variable}
\SetKwInput{Variables}{Local variables}
\SetKwInput{Function}{Local function}
\SetKwInput{Functions}{Local functions}
\SetKw{Let}{Let}
\SetKw{Add}{Add}
\SetKwComment{Comment}{{// }}{}

\usepackage{amsmath}
\DeclareMathOperator*{\argmax}{argmax}
\definecolor{fgreen}{rgb}{0.0, 0.5, 0.0}

\newcommand{\KGT}[3]{\ensuremath{{\langle}\texttt{#1},}
  \ensuremath{#2,} \ensuremath{\texttt{#3}{\rangle}}}
  
\newcommand{\GNODE}[1]{\texttt{#1}}
\newcommand{\GEDGE}[1]{\ensuremath{#1}}
\newcommand{\myNum}[1]{(\emph{#1})}

\def\OURMODEL{$\mathsf{RIDE}$}
\def\WITHPU{~\mathpzc{/w}~\textsc{PU}}
\def\NOPU{~\mathpzc{w/o}~\textsc{PU}}

\usepackage{layouts}
\usepackage{svg}
\usepackage{soul}

\usepackage[hang,flushmargin]{footmisc}
\newcommand{\bigA} {\mathcal{A} }

\newcommand{\bigI} {\mathcal{I} }
\newcommand{\bigG} {\mathcal{G} }

\newcommand{\bigO} {\mathcal{O} }

\newcommand{\bigR} {\mathcal{R} }
\newcommand{\bigS} {\mathcal{S} }
\newcommand{\bigU} {\mathcal{U} }
\newcommand{\bigX} {\mathcal{X} }
\newcommand{\bigY} {\mathcal{Y} }

\newcommand{\one} {\mathpzc{1} }
\newcommand{\two} {\mathpzc{2} }

\newcommand{\smalld} {\mathpzc{dim} }
\newcommand{\forward} {\mathpzc{fw} }
\newcommand{\backward} {\mathpzc{bw} }
\newcommand{\smallf} {f}
\newcommand{\smallg} {g}
\newcommand{\smallh} {\mathpzc{\textbf{h}}}
\newcommand{\smalli} {\mathpzc{i} }
\newcommand{\smallj} {\mathpzc{j} }
\newcommand{\smallk} {\mathpzc{k} }

\newcommand{\smallm} {\mathpzc{m} }

\newcommand{\smallq} {\mathpzc{q} }
\newcommand{\smallr} {\mathpzc{r} }
\newcommand{\smallt} {\mathpzc{t} }
\newcommand{\smallu} {\mathpzc{u} }

\newcommand{\smallw} {\mathpzc{w} }

\newcommand{\smalllp} {LP}
\newcommand{\relationship} {\mathpzc{relationship} }

\newcommand{\intent} {\mathpzc{intent} }

\newcommand{\utterance} {\mathpzc{utterance} }
\newcommand{\dimension} {\mathpzc{d} }
\newcommand{\emb} {\mathpzc{emb} }

\newcommand{\realR} {\mathbb{R} }

\title{Generalized Zero-shot Intent Detection via Commonsense Knowledge}

\author{A.B. Siddique, Fuad Jamour, Luxun Xu, Vagelis Hristidis
 \\
  University of California, Riverside \\
  \texttt{msidd005,fuadj,lxu051\{@ucr.edu\}, vagelis@cs.ucr.edu } }

\begin{document}
\maketitle
\begin{abstract}
Identifying user intents from natural language utterances is a crucial step in conversational systems that has been extensively studied as a supervised classification problem.
However, in practice, new intents emerge after deploying an intent detection model.
Thus, these models should seamlessly adapt and classify utterances with both seen and unseen intents -- unseen intents emerge after deployment and they do not have training data.
The few existing models that target this setting rely heavily on the scarcely available training data and overfit to seen intents data, resulting in a bias to misclassify utterances with unseen intents into seen ones.
We propose {\OURMODEL}: an intent detection model that leverages commonsense knowledge in an unsupervised fashion to overcome the issue of training data scarcity.
{\OURMODEL} computes robust and generalizable \emph{relationship meta-features} that capture deep semantic relationships between utterances and intent labels; these features are computed by considering how the concepts in an utterance are linked to those in an intent label via commonsense knowledge.
Our extensive experimental analysis on three widely-used intent detection benchmarks show that relationship meta-features significantly increase the accuracy of detecting both seen and unseen intents and that {\OURMODEL} outperforms the state-of-the-art model for unseen intents.
\end{abstract}

\input{sections/intro}
\input{sections/background}

\input{sections/model}
\input{sections/experiments}
\input{sections/results}
\input{sections/related}
\input{sections/conclusion}

\bibliography{anthology,custom}
\bibliographystyle{acl_natbib}

\appendix

\appendix
\balance
\input{sections/appendix}
\end{document}

%% file: sections/intro.tex
\section{Introduction}
\label{intro}
Virtual assistants such as Amazon Alexa and Google Assistant allow users to perform a variety of tasks (referred to as `skills' in Alexa) through an intuitive natural language interface.
For example, a user can set an alarm by simply issuing the utterance \emph{``Wake me up tomorrow at 10 AM''}
to a virtual assistant, and the assistant is expected to understand that the user's intent (i.e., ``AddAlarm'') is to invoke the alarm module, then set the requested alarm accordingly.
Detecting the intent implied in a natural language utterance  (i.e., \emph{intent detection}) is typically the first step towards performing any task in conversational systems.

Intent detection (or classification) is a challenging task due to the vast diversity in user utterances.
The challenge is further exacerbated in the more practically relevant setting where the full list of possible intents (or classes) is not available before deploying the conversational system, or intents are added over time.
This setting is an instance of the \emph{generalized zero-shot classification problem}~\cite{felix2018multi}:
labeled training utterances are available for seen intents but are unavailable for unseen ones, and at inference time, models do not have prior knowledge on whether the utterances they receive imply seen or unseen intents; i.e., unseen intents emerge after deploying the model.
This setting is the focus of this paper.

Little research has been conducted on building generalized zero-shot (GZS) models for intent detection, with little success.
The authors in~\cite{liu2019reconstructing} proposed a dimensional attention mechanism into a capsule neural network~\cite{sabour2017dynamic} and computing transformation matrices for unseen intents to accommodate the GZS setting.
This model overfits to seen classes and exhibits a strong bias towards classifying utterances into seen intents, resulting in poor performance.
Most recently, the authors in~\cite{yan2020unknown} extended the previous model by utilizing the density-based outlier detection algorithm LOF~\cite{breunig2000lof}, which allows distinguishing utterances with seen intents from those with unseen ones, which partially mitigates the overfitting issue. Unfortunately, the performance of this model is sensitive to that of LOF, which fails in cases where intent labels are semantically close. 

We propose {\OURMODEL\footnote{{\OURMODEL}: \underline{R}elationship Meta-features Assisted \underline{I}ntent \underline{DE}tection}}\footnote{Currently under review. GitHub Code Repository will be shared upon acceptance},
a model for GZS intent detection that utilizes commonsense knowledge to compute robust and generalizable unsupervised \emph{relationship meta-features}. 
These meta-features capture deep semantic associations between an utterance and an intent, resulting in two advantages:
\myNum{i} they significantly decrease the bias towards seen intents as they are similarly computed for both seen and unseen intents and \myNum{ii} they infuse commonsense knowledge into our model, which significantly reduces its reliance on training data without jeopardizing its ability to distinguish semantically close intent labels.
Relationship meta-features are computed by analyzing how the phrases in an utterance are linked to an intent label via commonsense knowledge.

\begin{figure}
    \centering
    \includegraphics[width=\linewidth]{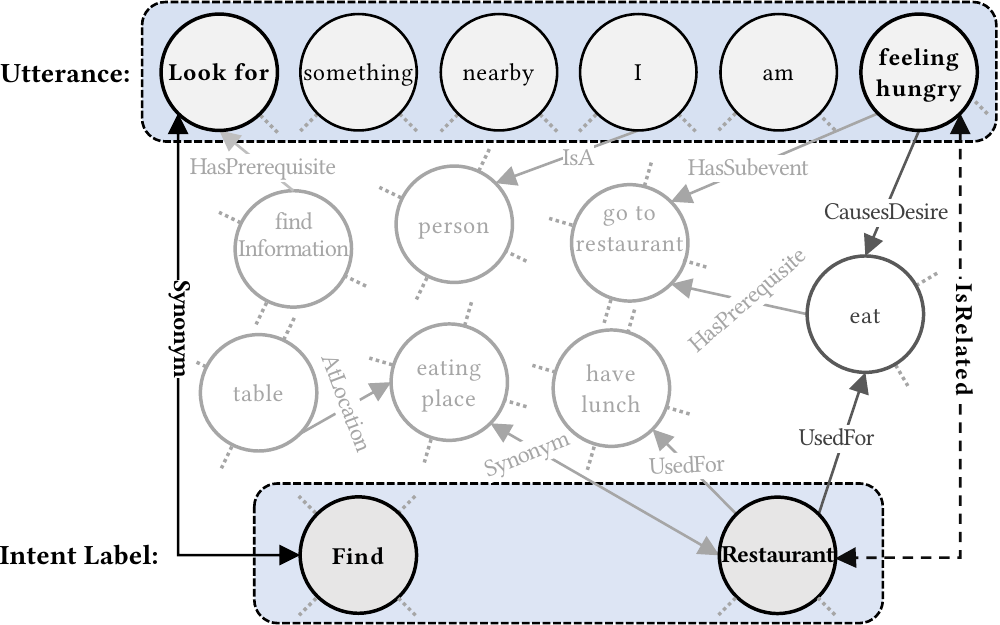}
    \caption{Example utterance, intent, and small commonsense knowledge graph. The presence of direct links such as \KGT{look for}{Synonym}{find} and inferred ones such as \KGT{feeling hungry}{IsRelated}{restaurant} between phrases in the utterance and those in the intent label indicate utterance-intent compatibility.}
    \label{fig:intro_fig}
    %\vspace{-10pt}
\end{figure}

Figure~\ref{fig:intro_fig} shows how the words (or phrases) in an example utterance are linked to the words in an example intent label through the nodes (i.e., concepts) of a commonsense knowledge graph.
In this example, the link \KGT{look for}{Synonym}{find} indicates that \GNODE{look for} and \GNODE{find} are synonyms, and the links \KGT{feeling hungry}{CausesDesire}{eat} and \KGT{restaurant}{UsedFor}{eat} can be used to infer the existence of the link \KGT{feeling hungry}{IsRelated}{restaurant}, which indicates that \GNODE{feeling hungry} and \GNODE{restaurant} are related. These two links, the direct and the inferred ones, carry a significant amount of semantic relatedness, which indicates that the given utterance-intent pair is compatible.
Note that this insight holds regardless of whether the intent is seen or not.
{\OURMODEL} utilizes this insight to build relationship meta-feature vectors that quantify the relatedness between an utterance and an intent in an unsupervised fashion.

{\OURMODEL} combines relationship meta-features with contextual word embeddings~\cite{Peters:2018}, and feeds the combined feature vectors into a trainable prediction function to finally detect intents in utterances.
Thanks to our relationship meta-features, {\OURMODEL} is able to accurately detect both seen and unseen intents in utterances.
Our extensive experimental analysis using the three widely used benchmarks, SNIPS~\cite{coucke2018snips},  SGD~\cite{rastogi2019towards}, and MultiWOZ~\cite{zang-etal-2020-multiwoz} show that our model outperforms the state-of-the-art model in detecting unseen intents in the GZS setting by at least 30.36\%.

A secondary contribution of this paper is that we managed to further increase the accuracy of GZS intent detection by employing Positive-Unlabeled (PU) learning~\cite{elkan2008learning} to predict if a new utterance belongs to a seen or unseen intent. PU learning assists intent detection models by mitigating their bias towards classifying most utterances into seen intents.
A PU classifier is able to perform binary classification after being trained using only positive and unlabeled examples. 
We found that the use of a PU classifier also improves the accuracy of existing GZS intent detection works, but our model is again outperforming these works.

%% file: sections/background.tex
\section{Preliminaries}
\label{sec:prelim}

\enlargethispage{20pt}

\begin{figure*}[t!]
  \centering
  \includegraphics[width=0.82\linewidth]{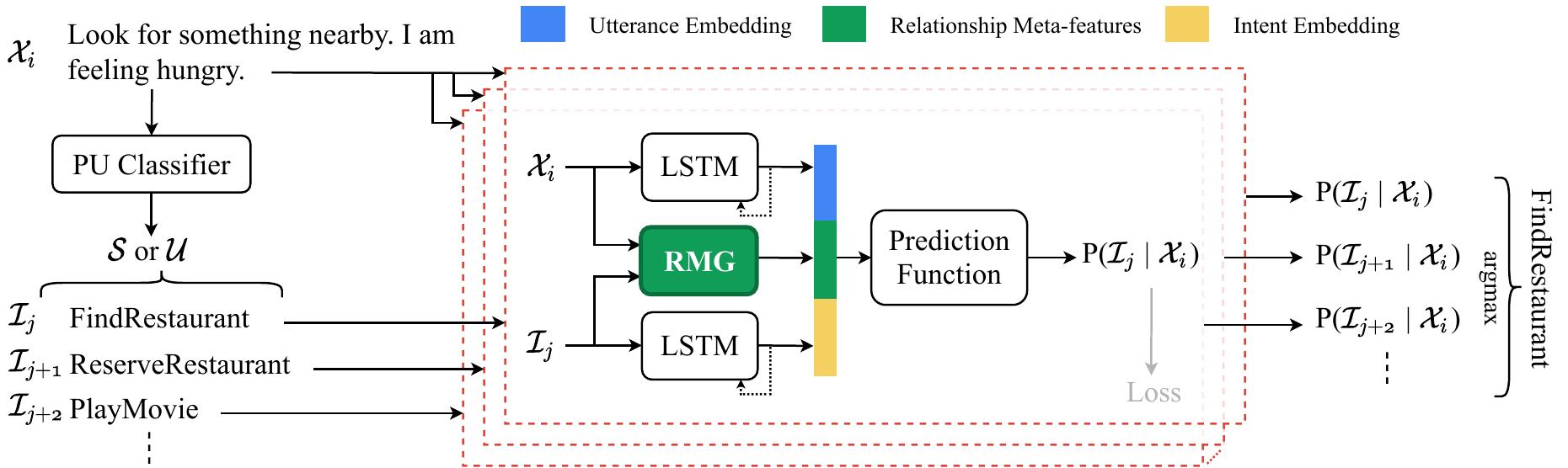}
  \caption{Overview of {\OURMODEL}.}
  \label{fig:gzs}
  %\vspace{-10pt}
\end{figure*}

\subsection{Intent Detection}
Let 
$\bigS = \{\bigI_\one,\cdots,\bigI_\smallk\}$ 
be a set of seen intents and
$\bigU = \{\bigI_{k+1}, \cdots,\bigI_n\}$ 
be a set of unseen intents where $\bigS\cap \bigU = \varnothing$.
Let 
$\bigX = \{\bigX_\one,\bigX_\two,...,\bigX_\smallm \}$ 
be a set of labeled training utterances where each training utterance $\bigX_i \in \bigX$ is described with a tuple $(\bigX_\smalli, \bigI_\smallj)$ such that $\bigI_\smallj \in \bigS$.
An intent $\bigI_\smallj$ is comprised of an \emph{Action} and an \emph{Object} and takes the form ``ActionObject''\footnote{If intents are described using a complex textual description, Actions and Objects can be extracted using existing NLP tools such as dependency parsers.} (e.g., ``FindRestaurant''); an Action describes a user's request or activity and an Object describes the entity pointed to by an Action~\cite{chen2013identifying,aaaiactobj,vedula2020open}.
In both the zero-shot (ZS) and the GZS settings, the training examples have intent labels from set $\bigS$ only, however, the two settings differ as follows.

\stitle{ZS Intent Detection.}
Given a test utterance $\bigX^\prime_i$ whose true label $\bigI_j$ is known to be in $\bigU$ a priori, predict a label $\bigI^\prime_\smallj \in \bigU$.

\stitle{GZS Intent Detection.}
Given a test utterance $\bigX^\prime_i$, predict a label $\bigI^\prime_\smallj \in \bigS \cup~ \bigU$. Note that unlike in the ZS setting, it is not known whether the true label of $\bigX^\prime_i$ belongs to $\bigS$ or $\bigU$, which exacerbates the challenge in this setting; we focus on this setting in this paper.

\subsection{Knowledge Graphs}
Knowledge graphs (KG) are structures that capture relationships between entities, and are typically used to capture knowledge in a semi-structured format; i.e., they are used as knowledge bases.
Knowledge graphs can be viewed as collections of triples, each representing a fact of the form \KGT{head}{relation}{tail} where \GNODE{head} and \GNODE{tail} describe entities and \GEDGE{relation} describes the relationship between the respective entities. In this work, we use ConceptNet~\cite{speer2016conceptnet}, which is a rich and widely-used commonsense knowledge graph. Interested readers can check Appendix~\ref{sec:kg_details} for more details on ConceptNet.

\subsection{Link Prediction}
While large knowledge graphs may capture a large subset of knowledge, they are incomplete: some relationships (or links) are missing.
Link prediction~\cite{kazemi2018simple} augments knowledge graphs by predicting missing relations using existing ones.
In the context of this work, we pre-train a state-of-the-art link prediction model~(LP) on the ConceptNet KG to score novel facts that are not necessarily present in the knowledge graph.
Given a triple (i.e., fact) in the form \KGT{head}{relation}{tail}, a link prediction model scores the triple with a value between 0 and 1, which quantifies the level of validity of the given triple.
The details of training our link predictor are available in Appendix~\ref{sec:app:lp}.

\subsection{Positive-Unlabeled Learning}
\label{pu_inference}
Positive-Unlabeled (PU) classifiers learn a standard binary classifier in the unconventional setting where labeled negative training examples are unavailable.
The state-of-the-art PU classifier~\cite{elkan2008learning}, which we integrate into our model, learns a decision boundary based on the positive and unlabeled examples, and thus can classify novel test examples into positive or negative. The aim of the PU classifier is to learn a probabilistic function $\smallf(\bigX_\smalli)$ that estimates $P(\bigI_\smallj \in \bigS \mid \bigX_\smalli)$ as closely as possible.
In this work, we train a PU classifier using our training set (utterances with only seen intents labeled as positive) and validation set (utterances with both seen and unseen intents as unlabeled). 
We use 512-dimensions sentence embedding as features when using the PU classifier, generated using a pre-trained universal sentence encoder~\cite{cer2018universal}.

%% file: sections/model.tex
\section{Our Approach}
\label{model}
Figure~\ref{fig:gzs} shows an overview of our model:
given an input utterance $\bigX_i$, we first invoke the PU classifier (if it is available) to predict whether $\bigX_i$ implies a seen or an unseen intent; i.e., whether $\bigX_i$'s intent belongs to set $\bigS$ or $\bigU$. Then, an instance of our core model (the red box in Figure~\ref{fig:gzs}) is invoked for each intent in $\bigS$ or $\bigU$ based on the PU's prediction.
Our core model predicts the level of compatibility between the given utterance $\bigX_i$ and intent $\bigI_j$, i.e., the probability that the given utterance implies the given intent $P(\bigI_j|\bigX_i) \in [0,1]$. Finally, our model outputs the intent with the highest compatibility probability, i.e., $\argmax_{\bigI_j} P(\bigI_j|\bigX_i)$.

Our core model concatenates relationship meta-features, utterance embedding, and intent embedding and feeds them into a trainable prediction function.
The Relationship Meta-features Generator (RMG) is at the heart of our model, and it is the most influential component. Given an utterance and an intent, RMG generates meta-features that capture deep semantic associations between the given utterance and intent in the form of a meta-feature vector.

\subsection{Relationship Meta-feature Generation}
\label{sec:rmg}
RMG extracts relationship meta-features by utilizing the ``ActionObject'' structure of intent labels and commonsense knowledge graphs. Relationship meta-features are not only generalizable, but also discriminative: while the example utterance ``Look for something nearby. I am feeling hungry.'' may be related to the intent ``ReserveRestaurant'', the Action part of this intent is not related to any phrase in the utterance; thus, ``ReserveRestaurant'' is less related to the utterance than ``FindRestaurant''. 

RMG takes the following inputs:
a set of relations in a knowledge graph (35 in the case of ConceptNet) $\bigR = \{ \smallr_\one, \smallr_\two,.., \smallr_\smallt \}$;
the set of n-grams $\bigG_\smalli = \{ \smallg_\one, \smallg_\two,.., \smallg_\smallq \}$ that correspond to the input utterance $\bigX_\smalli$, where $|\bigG| = \smallq$;
and an intent label $\bigI_\smallj = \{ \bigA, \bigO \}$, where $\bigA$ and $\bigO$ are the Action and Object components of the intent, respectively.
RMG computes a relationship meta-features vector in four steps, where each step results in a vector of size $|\bigR|$. The smaller vectors are: 
$\textbf{e}^{\overrightarrow{\bigA}}_{\bigX_\smalli}$, 
$\textbf{e}^{\overrightarrow{\bigO}}_{\bigX_\smalli}$, 
$\textbf{e}^{\overleftarrow{\bigA}}_{\bigX_\smalli}$, and 
$\textbf{e}^{\overleftarrow{\bigO}}_{\bigX_\smalli}$,
where $\textbf{e}^{\overrightarrow{\bigA}}_{\bigX_\smalli}$ captures the Action to utterance semantic relationships and $\textbf{e}^{\overrightarrow{\bigO}}_{\bigX_\smalli}$ captures the Object to utterance relationships. The remaining two vectors capture relationships in the other direction; i.e., utterance to Action/Object, respectively. Capturing bi-directional relationships is important because a relationship in one direction does not necessarily imply one in the other direction -- for example, \KGT{table}{AtLocation}{restaurant} does not imply \KGT{restaurant}{AtLocation}{table}. The final output of RMG is the relationship meta-features vector $\textbf{e}_{\relationship}$, which is the concatenation of the four aforementioned vectors. We explain next how the smaller vectors is computed.

\setlength{\algomargin}{1pt}
\begin{algorithm}[t!]
\footnotesize
\DontPrintSemicolon
\SetKwFunction{RM}{RM}
\SetKwFunction{Max}{Max}
  \KwInput{
  \hspace{-3pt}$\bigR = \{ \smallr_\one, \cdots, \smallr_\smallt \}$: relations in KG \\
  \hspace{24pt}$\bigG_i = \{ \smallg_\one, \cdots, \smallg_\smallq \}$: utterance n-grams\\
  \hspace{24pt}$\bigI_j = \{ \bigA, \bigO \}$: intent's Action and Object}
  \vspace{5pt}
  \KwOutput{$\textbf{e}_{\relationship}$: $\bigX_i$-$\bigI_j$ relationship meta-features}
%   \vspace{3pt}
Let $\textbf{e}^{\overrightarrow{\bigA}}_{\bigX_\smalli}$ = \RM($\bigA$, $\bigG_\smalli$, $\rightarrow$)  // Action to utterance \;
Let $\textbf{e}^{\overrightarrow{\bigO}}_{\bigX_\smalli}$ = \RM($\bigO$, $\bigG_\smalli$, $\rightarrow$) // Object to utterance \;
Let $\textbf{e}^{\overleftarrow{\bigA}}_{\bigX_\smalli}$  = \RM($\bigA$, $\bigG_\smalli$, $\leftarrow$) // utterance to Action\; 
Let $\textbf{e}^{\overleftarrow{\bigO}}_{\bigX_\smalli}$  = \RM($\bigO$, $\bigG_\smalli$, $\leftarrow$) // utterance to Object\;
Let $\textbf{e}_{\relationship}$ = [$\textbf{e}^{\overrightarrow{\bigA}}_{\bigX_\smalli}$, $\textbf{e}^{\overrightarrow{\bigO}}_{\bigX_\smalli}$, $\textbf{e}^{\overleftarrow{\bigA}}_{\bigX_\smalli}$, $\textbf{e}^{\overleftarrow{\bigO}}_{\bigX_\smalli}$] \;
\BlankLine
\KwRet  $\textbf{e}_{\relationship}$ \;
\BlankLine
% \BlankLine
% \BlankLine
  \SetKwProg{Fn}{Function}{:}{}
  \Fn{\RM{$concept$, $phrases$, $direction$}}{
        Let $\textbf{e}$ = [] \;
        \ForEach{$ r \in \bigR$}
        {
            \If{$direction$ = $\rightarrow$}
            {
                 Let $p$ = \Max$(\smalllp(concept, r, g))$ for $g \in phrases$ \;
            }
            \If{$direction$ = $\leftarrow$}
            {
                 Let $p$ = \Max$(\smalllp(g, r, concept))$ for $g \in phrases$ \;
            }
            $\textbf{e}$.append(p)
        }
        \KwRet $\textbf{e}$\;
  }
\caption{RMG}
\label{algo:reg}
\end{algorithm}

RMG computes $\textbf{e}^{\overrightarrow{\bigA}}_{\bigX_\smalli}$ by considering the strength of each relation in $\bigR$ between $\bigA$ and each n-gram in $\bigG_\smalli$. That is, $\textbf{e}^{\overrightarrow{\bigA}}_{\bigX_\smalli}$ has $|\bigR|$ cells, where each cell corresponds to a relation $r \in \bigR$. Each cell is computed by taking $max(LP(\bigA, r, g))$ over all $g \in \bigG_\smalli$. 
$LP$(\GNODE{head}, \GEDGE{relation}, \GNODE{tail}) outputs the probability that the fact represented by the triple \KGT{head}{relation}{tail} exists. 
The vector $\textbf{e}^{\overrightarrow{\bigO}}_{\bigX_\smalli}$ is computed similarly, but with passing $\bigO$ instead of $\bigA$ when invoking the link predictor; i.e., taking $max(LP(\bigO, r, g))$ over all $g \in \bigG_\smalli$ to compute each cell. The vectors $\textbf{e}^{\overleftarrow{\bigA}}_{\bigX_\smalli}$ and 
$\textbf{e}^{\overleftarrow{\bigO}}_{\bigX_\smalli}$ are computed similarly, but with swapping the \GNODE{head} and \GNODE{tail} when invoking the link predictor; i.e., utterance phrases are passed as \GNODE{head} and Action/Object parts are passed as \GNODE{tail}.
Algorithm~\ref{algo:reg} outlines the previous process.
Finally, the generated meta-features are passed through a linear layer with sigmoid activation before concatenation with the utterance and intent embeddings.

\subsection{Utterance and Intent Encoders}
\label{gru}
Given an utterance $\bigX_\smalli = \{\smallw_\one, \smallw_\two, \cdots, \smallw_{\smallu} \}$ with $\smallu$ words,
 first we compute an embedding $\emb(\smallw_\smallj) \in \realR^{\smalld}$ for each word $\smallw_{\smallj}$ in the utterance, where $\emb(\smallw_\smallj)$ is the concatenation of a contextual embedding obtained from a pre-trained ELMo model and parts of speech (POS) tag embedding. Then, we use bi-directional LSTM to produce a $\dimension$-dimensional representation as follows:
\begin{equation*}
  \overrightarrow{\smallh}_\smallj = \text{LSTM}_\forward(\overrightarrow{\smallh}_{\smallj-\one}, \emb(\smallw_\smallj)).
  \label{eq:encoder-fw}
\end{equation*}
\begin{equation*}
  \overleftarrow{\smallh}_\smallj = \text{LSTM}_\backward(\overleftarrow{\smallh}_{\smallj-\one}, \emb(\smallw_\smallj)).
  \label{eq:encoder-bw}
\end{equation*}
Finally, we concatenate the output of the last hidden states as utterance embedding $\textbf{e}_{\utterance} = [\overrightarrow{\smallh}_\smallu ; \overleftarrow{\smallh}_\smallu] \in \realR^{\dimension}$.
We encode intent labels similarly to produce an intent embedding  $\textbf{e}_{\intent} \in \realR^{\dimension}$.

\subsection{Training the Model}
\label{training}
Our model has two trainable components: the LSTM units in the utterance and intent encoders and the compatibility probability prediction function.
We jointly train these components using training data prepared as follows.
The training examples are of the form $((\bigX_\smalli, \bigI_\smallj), \bigY)$, where $\bigY$ is a binary label representing whether the utterance-intent pair $(\bigX_\smalli, \bigI_\smallj)$ are compatible: $1$ means they are compatible, and $0$ means they are not. For example, the utterance-intent pair (``I want to play this song'', ``PlaySong'') gets a label of $1$, and the same utterance paired with another intent such as ``BookHotel'' gets a label of $0$.
We prepare our training data by assigning a label of $1$ to the available utterance-intent pairs (where intents are seen ones); these constitute positive training examples. We create a negative training example for each positive one by corrupting the example's intent. We corrupt intents by modifying their Action, Object, or both; for example, the utterance-intent pair (``Look for something nearby. I am hungry.'', ``FindRestaurant'') may result in the negative examples (..., ``ReserveRestaurant''), (..., ``FindHotel''), or (...,``RentMovies'').
We train our core model by minimizing the cross-entropy loss over all the training examples.

%% file: sections/experiments.tex
\section{Experimental Setup}
\label{experiments}
In this section, we describe the datasets, evaluation settings and  metrics, competing methods, and implementation details of our proposed method.

\subsection{Datasets}
\label{datasets}
Table~\ref{tab:dataset} presents important statistics on the datasets we used in our experiments.

\stitle{SNIPS}~\cite{coucke2018snips}.
A crowd-sourced single-turn NLU benchmark with $7$ intents across different domains.

\stitle{SGD}~\cite{rastogi2019towards}.
A recently published dataset for the eighth Dialog System Technology Challenge, Schema Guided Dialogue (SGD) track. It contains dialogues from $16$ domains with a total of $46$ intents. It is one of the most comprehensive and challenging publicly available datasets.
The dialogues contain user intents as part of dialogue states. We only kept utterances where users express an intent by comparing two consecutive dialogue states to check for the expression of a new intent.

\stitle{MultiWOZ}~\cite{zang-etal-2020-multiwoz}.
Multi-Domain Wizard-of-Oz (MultiWOZ) is a well-known and publicly available dataset. We used the most recent version $2.2$ of MultiWOZ in our experiments, which contains utterances spanning $11$ intents. Similarly to the pre-processing of SGD dataset, we only kept utterances that express an intent to maintain consistency with the previous work.

\begin{table}[t!]
\footnotesize
\centering
\begin{tabular}{lccc}
\hline
\textbf{Dataset}         & \textbf{SNIPS} & \textbf{SGD} & \textbf{MultiWOZ} 
\\ \hline
Dataset Size    & $14.2$K     & $57.2$K & $30.0$K       \\
Vocab. Size & $10.8$K     & $8.8$K  & $9.7$K    \\
Avg. Length & $9.05$ & $10.62$ & $11.07$ \\
\# of Intents    & $7$         & $46$ & $11$        \\
\hline
\end{tabular}
\caption{Dataset statistics.}
%\vspace{-10pt}
\label{tab:dataset}

\end{table}

\subsection{Evaluation Methodology}
\label{testing}
We use standard classification evaluation measures: accuracy and F1 score. The values for all the metrics are per class averages weighted by their respective support. We present evaluation results for the following intent detection settings:

\stitle{ZS intent Detection.}
In this setting, a model is trained on all the utterances with seen intents -- i.e., all samples $(\bigX_\smalli, \bigI_\smallj)$ where $\bigI_\smallj \in \bigS$.
Whereas at inference time, the utterances are only drawn from those with unseen intents; the model has to classify a given utterance into one of the unseen intents.
Note that this setting is less challenging than the GZS setting because models know that utterances received at inference time imply intents that belong to the set of unseen intents only, thus naturally reducing their bias towards classifying utterances into seen intents.
For each dataset, we randomly place $\approx 25\%$, $\approx 50\%$, and $\approx 75\%$ of the intents in the seen set for training and the rest into the unseen set for testing, and report the average results over $10$ runs.
It is important to highlight that selecting seen/unseen sets in this fashion is more challenging to models because all the intents get an equal chance to appear in the unseen set, which exposes models that are capable of detecting certain unseen intents only.

\stitle{GZS intent Detection.}
In this setting, models are trained on a subset of utterances implying seen intents. At inference time, test utterances are drawn from a set that contains utterances implying a mix of seen and unseen intents (disjoint set from training set) and the model is expected to select the correct intent from all seen and unseen intents for a given test utterance. This is the most realistic and challenging problem setting, and it is the main focus of this work. For the GZS setting, we decided the train/test splits for each dataset as follows:
For SNIPS, we first randomly selected 5 out of 7 intents and designated them as seen intents -- the remaining 2 intents were designated as unseen intents. We then selected $70\%$ of the utterances that imply any of the 5 seen intents for training. The test set consists of the remaining $30\%$ utterances in addition to all utterances that imply one of the 2 unseen intents.
Previous work~\cite{liu2019reconstructing} used the same number of seen/unseen intents, but selected the seen/unseen intents manually. Whereas we picked unseen intents randomly, and we report results over $10$ runs resulting in a more challenging and thorough evaluation. That is, each intent gets an equal chance to appear as an unseen intent in our experiments, which allows testing each model more comprehensively.
For SGD, we used the standard splits proposed by the dataset authors. Specifically, the test set includes utterances that imply $8$ unseen intents and $26$ seen intents; we report average results over $10$ runs.
For MultiWOZ, we used $70\%$ of the utterance that imply $8$ (out of $11$) randomly selected intents for training and the rest of the utterances (i.e., the remaining $30\%$ of seen intents' utterances and all utterances implying unseen intents) for testing.

\begin{figure*}[t]
    \centering
    \begin{subfigure}[b]{0.32\textwidth}
        \centering
        \includegraphics[width=\linewidth]{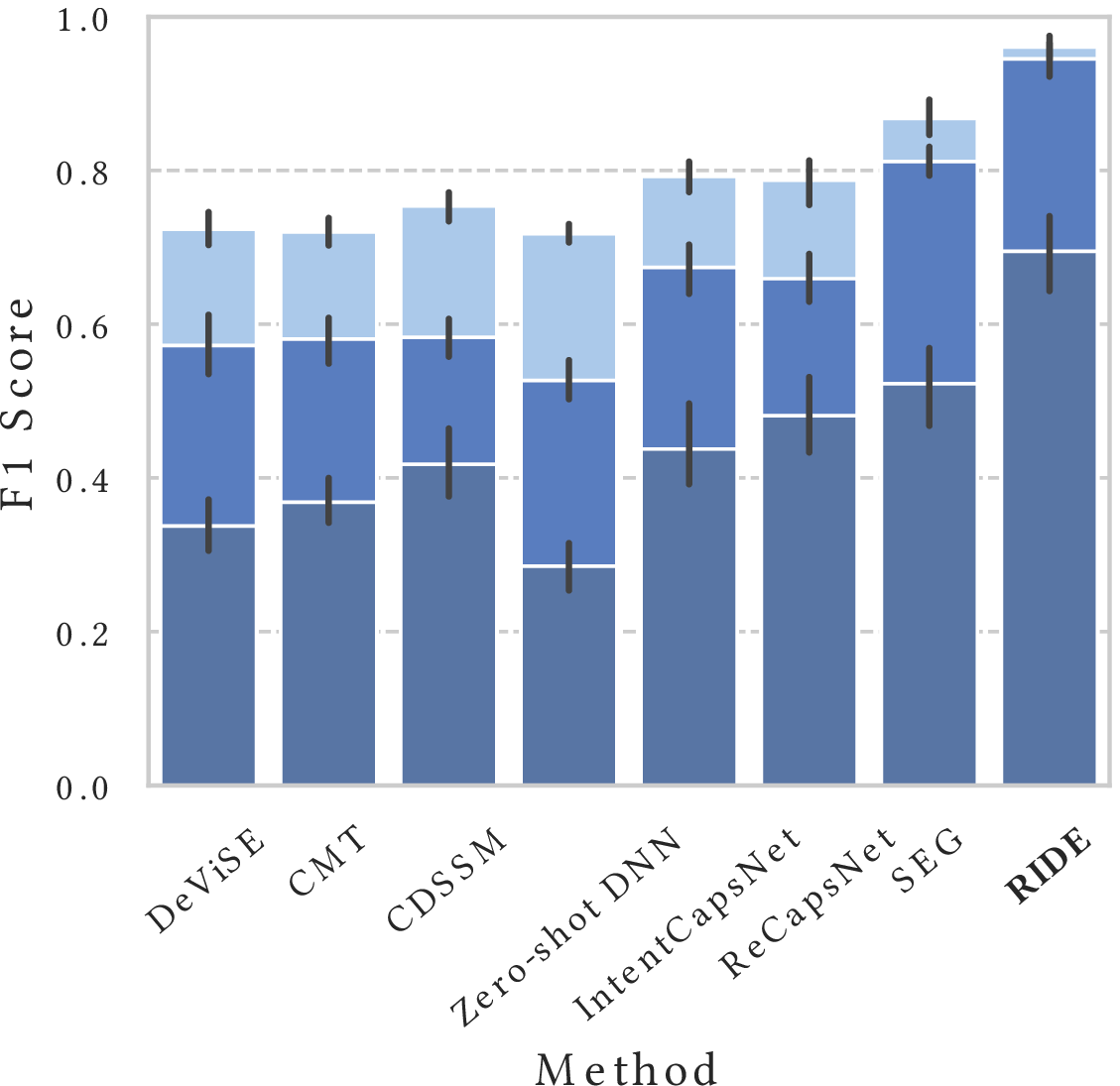}
        \caption{SNIPS dataset}
    \end{subfigure}  
    \hfill
    \begin{subfigure}[b]{0.32\textwidth}
        \centering
        \includegraphics[width=\linewidth]{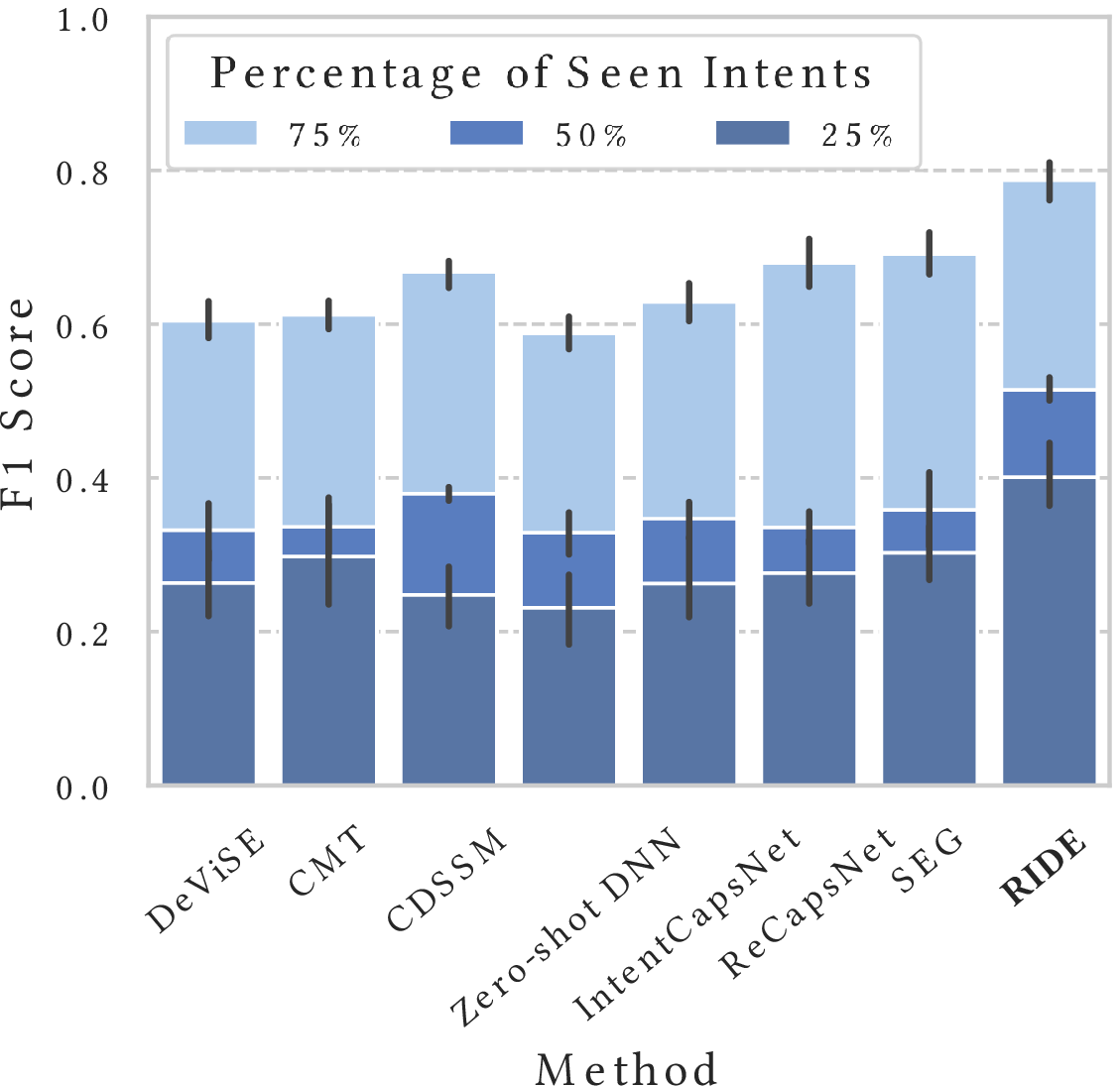}
        \caption{SGD dataset}
    \end{subfigure} 
    \hfill
    \begin{subfigure}[b]{0.32\textwidth}
        \centering
        \includegraphics[width=\linewidth]{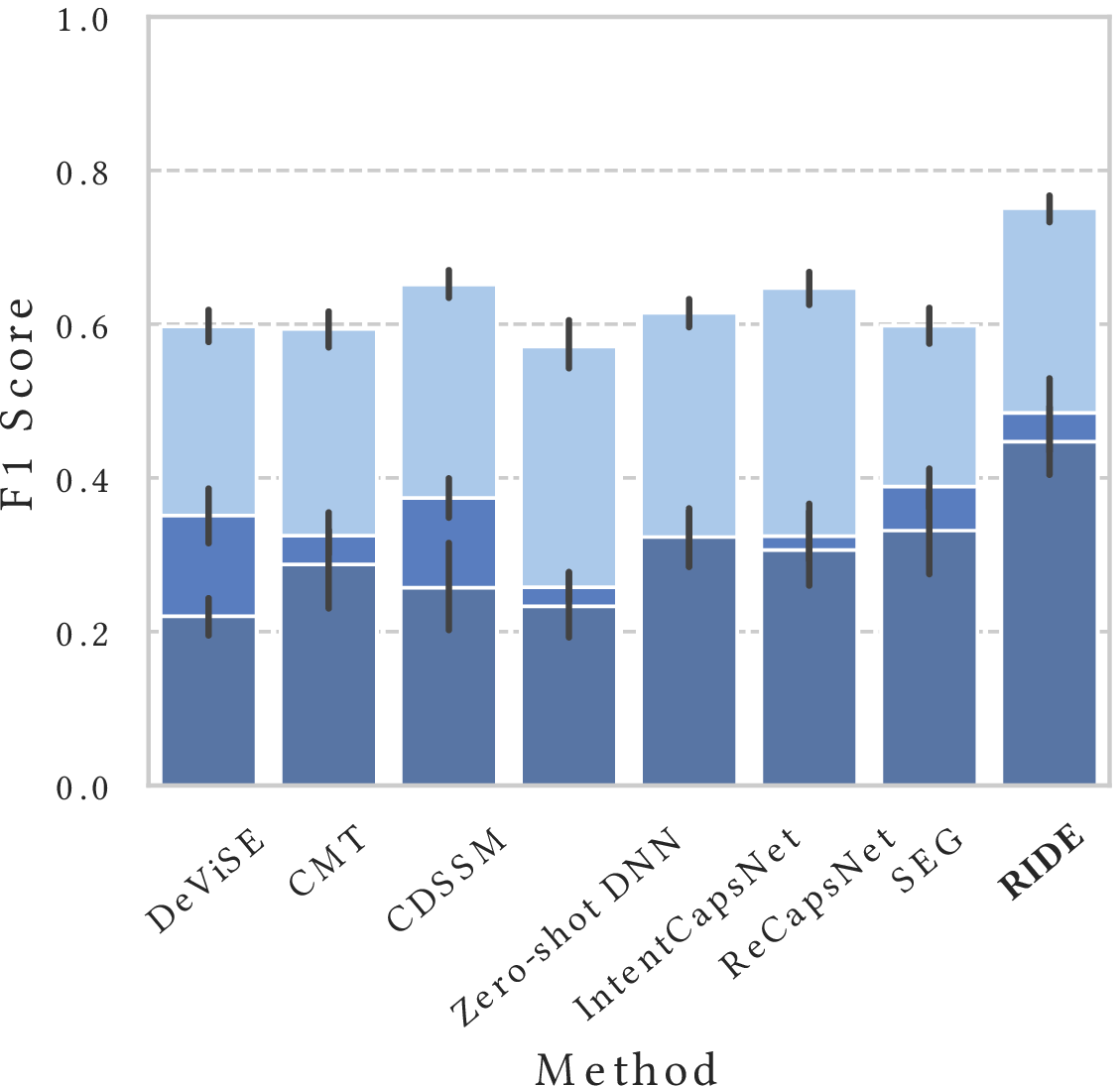}
        \caption{MultiWOZ dataset}
    \end{subfigure} 
    \caption{F1 scores for competing models in the ZS setting with varying percentages of seen intents. In the ZS setting, utterances with only unseen intents are encountered at inference time, and models are aware of this.
    Our model {\OURMODEL} consistently outperforms all other models for any given percentage of seen intents.
    }
\label{fig:zs}
%\vspace{-4pt}
\end{figure*}
\subsection{Competing Methods}
\label{baselines}

We compare our model {\OURMODEL} against the following state-of-the-art (SOTA) models and several strong baselines:

\stitle{SEG}~\cite{yan2020unknown}.
A semantic-enhanced Gaussian mixture model that uses large margin loss to learn class-concentrated embeddings coupled with a density-based outlier detection algorithm LOF to detect unseen intents. 

\stitle{ReCapsNet-ZS}~\cite{liu2019reconstructing}.
A model that employs capsule neural network and a dimensional attention module to learn generalizable transformational metrices from seen intents.

\stitle{IntentCapsNet}~\cite{xia2018zero}.
A model that utilizes capsule neural networks to learn low-level features and routing-by-agreement to adapt to unseen intents. This model was originally proposed for detecting intents in the standard ZS setting. We extended it to support the GZS setting with the help of its authors.

\stitle{Other Baseline Models}.
\myNum{i}~Zero-shot DDN~\cite{kumar2017zero}:
A model for ZS intent detection that achieves zero-shot capabilities by projecting utterances and intent labels into the same high dimensional embedding space.
\myNum{ii}~CDSSM~\cite{chen2016zero}:
A model for ZS intent detection that utilizes a convolutional deep structured semantic model to generate embeddings for unseen intents. 
\myNum{iii}~CMT~\cite{socher2013zero}:
A model for ZS intent detection that employs non-linearity in the compatibility function between utterances and intents to find the most compatible unseen intents. 
\myNum{iv}~DeViSE~\cite{frome2013devise}:
A model that was originally proposed for zero-shot image classification that learns a linear compatibility function.
Note that baseline ZS models have been extended to support GZS setting.

\subsection{Implementation Details}
\label{implementation}
We lemmatize ConeptNet KG, that has $1$ million nodes (English only after lemmatization), $2.7$ million edges, and $35$ relation types.
The link predictor is trained on the lemmatized version of ConceptNet KG.
The link predictor has two $200$-dimensional embedding layers and a negative sampling ratio of $10$; it is trained for $1,000$ epochs using Adam optimizer with a learning rate of $0.05$, L2 regularization value of $0.1$, and batch size of $4800$.
Our relationship meta-features generator takes in an utterance's n-grams with n $\le 4$ and an intent label, and uses the pre-trained link predictor to produce relationship meta-features with $140$ dimensions. 
Our utterance and intent encoders use pre-trained ELMo contextual word embeddings with $1024$ dimension and POS tags embeddings with $300$ dimension, and a two-layer RNN with $300$-dimensional bidirectional LSTM as recurrent units.
Our prediction function has two dense layers with relu and softmax activation.
Our core model is trained for up to $200$ epochs with early stopping using Adam optimizer and a cross entropy loss with initial learning rate of $0.001$ and ReduceLROnPlateau scheduler~\cite{torchopt68:online} with $20$ patience epochs. It uses dropout rate of $0.3$ and batch size of $32$. A negative sampling ratio of up to $6$ is used. We use the same embeddings generation and training mechanism for all competing models. 

%% file: sections/results.tex
\begin{table*}[t!]
\scriptsize
\centering
\begin{tabular}{l|ll|ll|ll|ll|llll}
\hline
\multirow{3}{*}{\textbf{Method}} & \multicolumn{4}{c|}{\textbf{SNIPS}}                                                                        & \multicolumn{4}{c|}{\textbf{SGD}}                                                                        & \multicolumn{4}{c}{\textbf{MultiWOZ}}                                                                            \\ \cline{2-13} 
                        & \multicolumn{2}{c|}{Unseen}                       & \multicolumn{2}{c|}{Seen}                         & \multicolumn{2}{c|}{Unseen}                       & \multicolumn{2}{c|}{Seen}                         & \multicolumn{2}{c|}{Unseen}                           & \multicolumn{2}{c}{Seen}                         \\ \cline{2-13} 
                        & \multicolumn{1}{c}{Acc} & \multicolumn{1}{c|}{F1} & \multicolumn{1}{c}{Acc} & \multicolumn{1}{c|}{F1} & \multicolumn{1}{c}{Acc} & \multicolumn{1}{c|}{F1} & \multicolumn{1}{c}{Acc} & \multicolumn{1}{c|}{F1} & \multicolumn{1}{c}{Acc} & \multicolumn{1}{c|}{F1}     & \multicolumn{1}{c}{Acc} & \multicolumn{1}{c}{F1} \\ \hline
DeViSE                  & 0.0311                  & 0.0439                  & 0.9487                  & 0.6521                  & 0.0197                  & 0.0177                  & 0.8390                  & 0.5451                  & 0.0119                  & \multicolumn{1}{l|}{0.0270} & 0.8980                  & 0.5770                 \\
CMT                     & 0.0427                  & 0.0910                 & \textbf{0.9751}                  & 0.6639                  & 0.0254                  & 0.0621                  & \textbf{0.9014}                  & 0.5803                  & 0.0253                  & \multicolumn{1}{l|}{0.0679} & \underline{0.9025}                  & 0.6216                 \\
CDSSM                   & 0.0521                  & 0.0484                  & 0.9542                  & 0.7028                  & 0.0367                  & 0.0284                  & 0.8890                  & 0.6379                  & 0.0373                  & \multicolumn{1}{l|}{0.0244} & 0.8861                  & 0.6515                 \\
Zero-shot DNN           & 0.0912                  & 0.1273                  & 0.9437                  & 0.6687                  & 0.0662                  & 0.1168                  & 0.8825                  & 0.6098                  & 0.0802                  & \multicolumn{1}{l|}{0.1149} & 0.8940                  & 0.6012                 \\
IntentCapsNet           & 0.0000                  & 0.0000                  & \underline{0.9749}                  & 0.6532                  & 0.0000                  & 0.0000                  & \underline{0.8982}                  & 0.5508                  & 0.0000                  & \multicolumn{1}{l|}{0.0000} & \textbf{0.9249}                  & 0.6038                 \\
ReCapsNet               & 0.1249                  & 0.1601                  & 0.9513                  & 0.6783                  & 0.1062                  & 0.1331                 & 0.8762                  & 0.5751                  & 0.1081                  & \multicolumn{1}{l|}{0.1467} & 0.8715                  & 0.6170                 \\
SEG                     & 0.6943                  & 0.6991                  & 0.8643                  & 0.8651                  & 0.3723                  & 0.4032                  & 0.6134                  & 0.6356                  & \underline{0.3712}                  & \multicolumn{1}{l|}{0.4143} & 0.6523                  & 0.6456                 \\ \hline
{\OURMODEL} ${\NOPU}$             & \underline{0.8728}                  &    \underline{0.9103}               & 0.8906                  &   \underline{0.8799}               & \underline{0.3865}                  & \underline{0.4634}                  & 0.8126                  &      \underline{0.8295}             & 0.3704                  & \multicolumn{1}{l|}{\underline{0.4645}} & 0.8558                  & \underline{0.8816}                 \\
{\OURMODEL} ${\WITHPU}$             & \textbf{0.9051}                  & \textbf{0.9254}                  & 0.9179                  & \textbf{0.9080}                   & \textbf{0.5901}                  & \textbf{0.5734}                  & 0.8315                  & \textbf{0.8298}                  & \textbf{0.5686}                  & \multicolumn{1}{l|}{\textbf{0.5206}} & 0.8844                  & \textbf{0.8847}      \\          
\hline
\end{tabular}
\caption{Main results: accuracy and F1 scores for competing models in the GZS setting (i.e., models receive both seen and unseen intents at inference time, which makes the setting more challenging than the ZS setting).
We present results for two variants of our model: {\OURMODEL}~${\NOPU}$ which does not use a PU classifier, and {\OURMODEL}~${\WITHPU}$ which uses one.
Our model consistently achieves the best F1 score for both seen and unseen intents across all datasets, regardless of whether the PU classifier is integrated or not.
}
\label{tab:generalized}
\end{table*}

\begin{figure*}[t]
    \centering
    \begin{subfigure}[b]{0.29\textwidth}
        \centering
        \includegraphics[width=\linewidth]{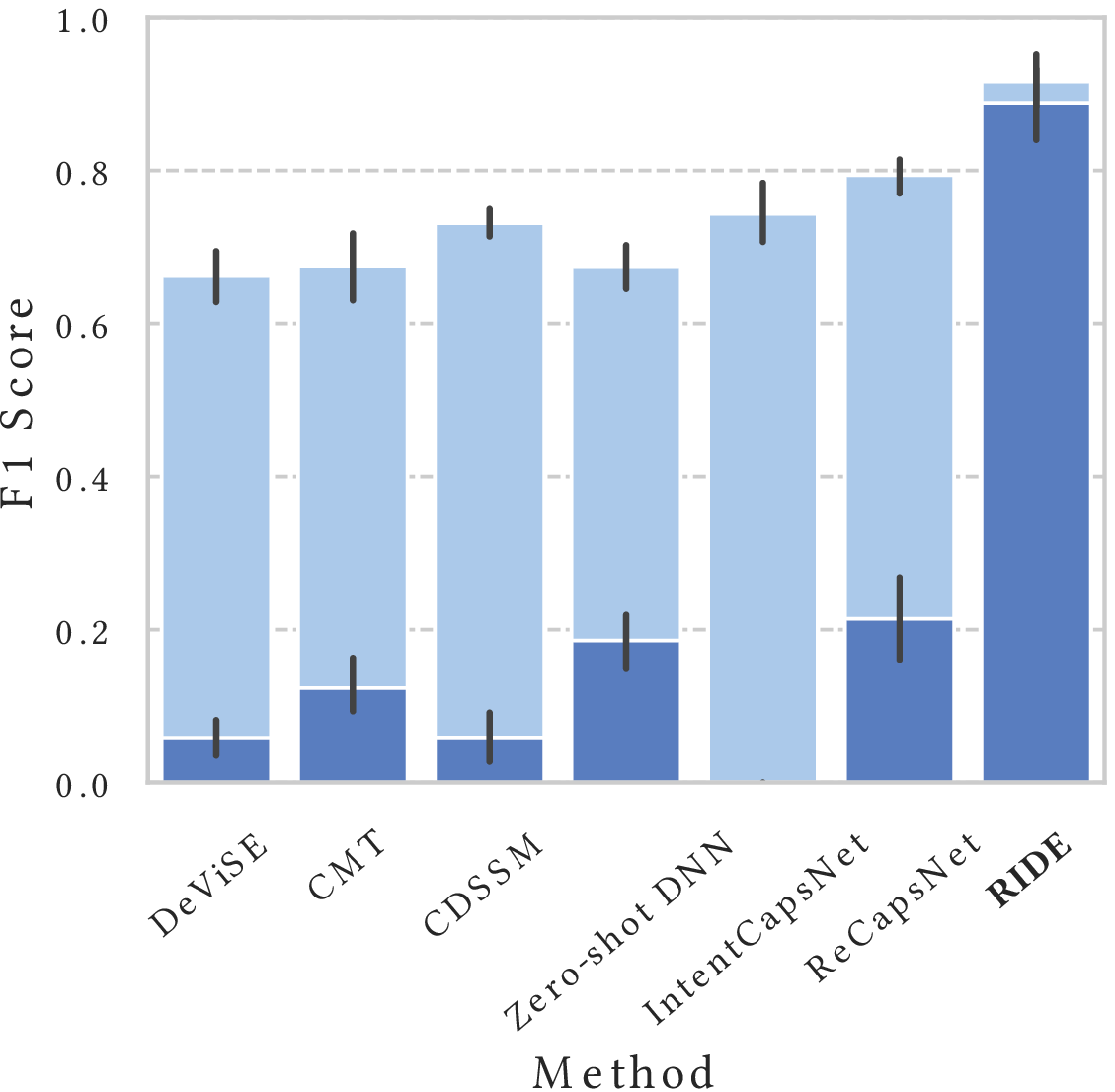}
        \caption{SNIPS dataset}
    \end{subfigure} 
    \hfill
    \begin{subfigure}[b]{0.29\textwidth}
        \centering
        \includegraphics[width=\linewidth]{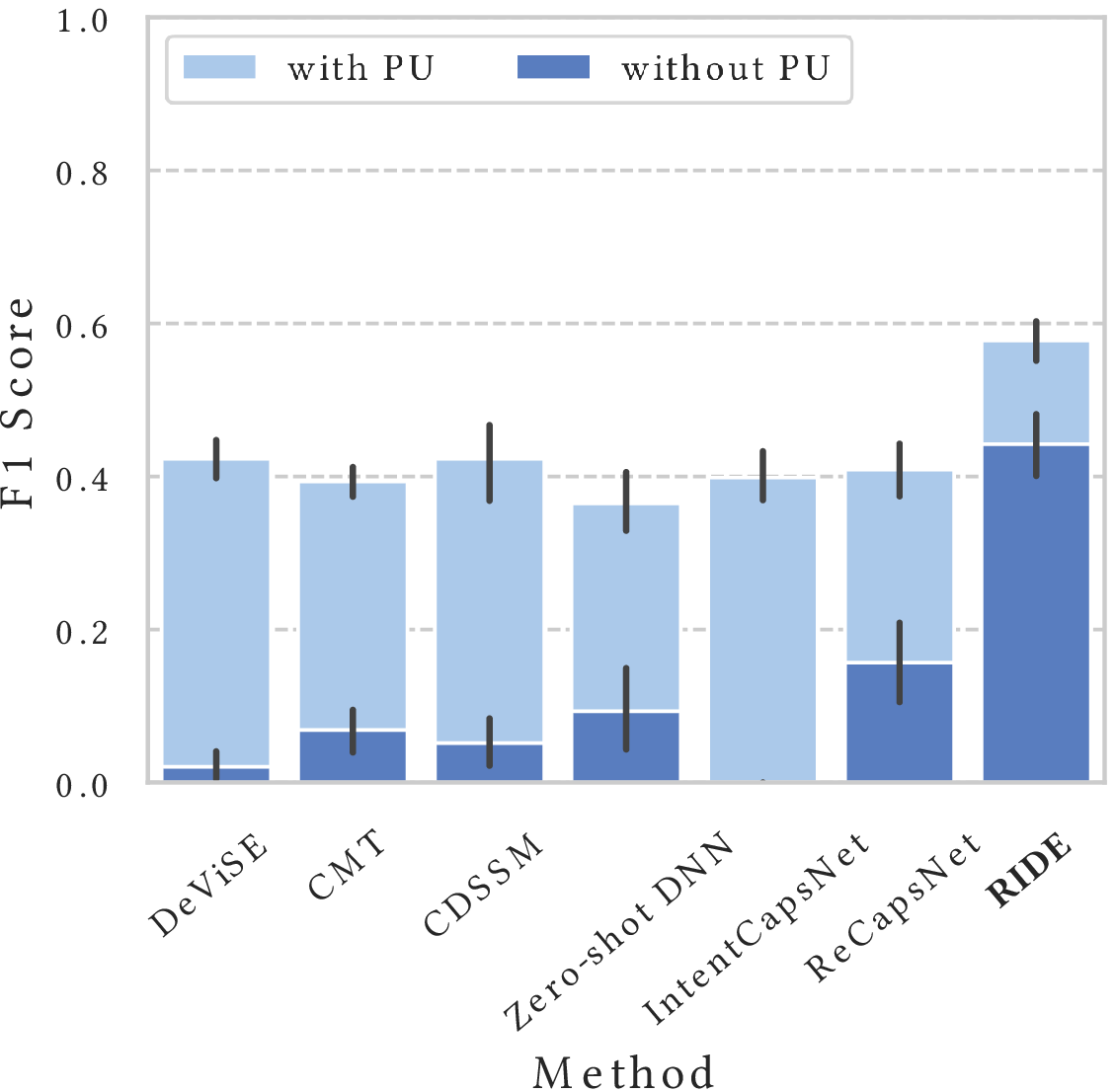}
        \caption{SGD dataset}
    \end{subfigure} 
    \hfill
    \begin{subfigure}[b]{0.29\textwidth}
        \centering
        \includegraphics[width=\linewidth]{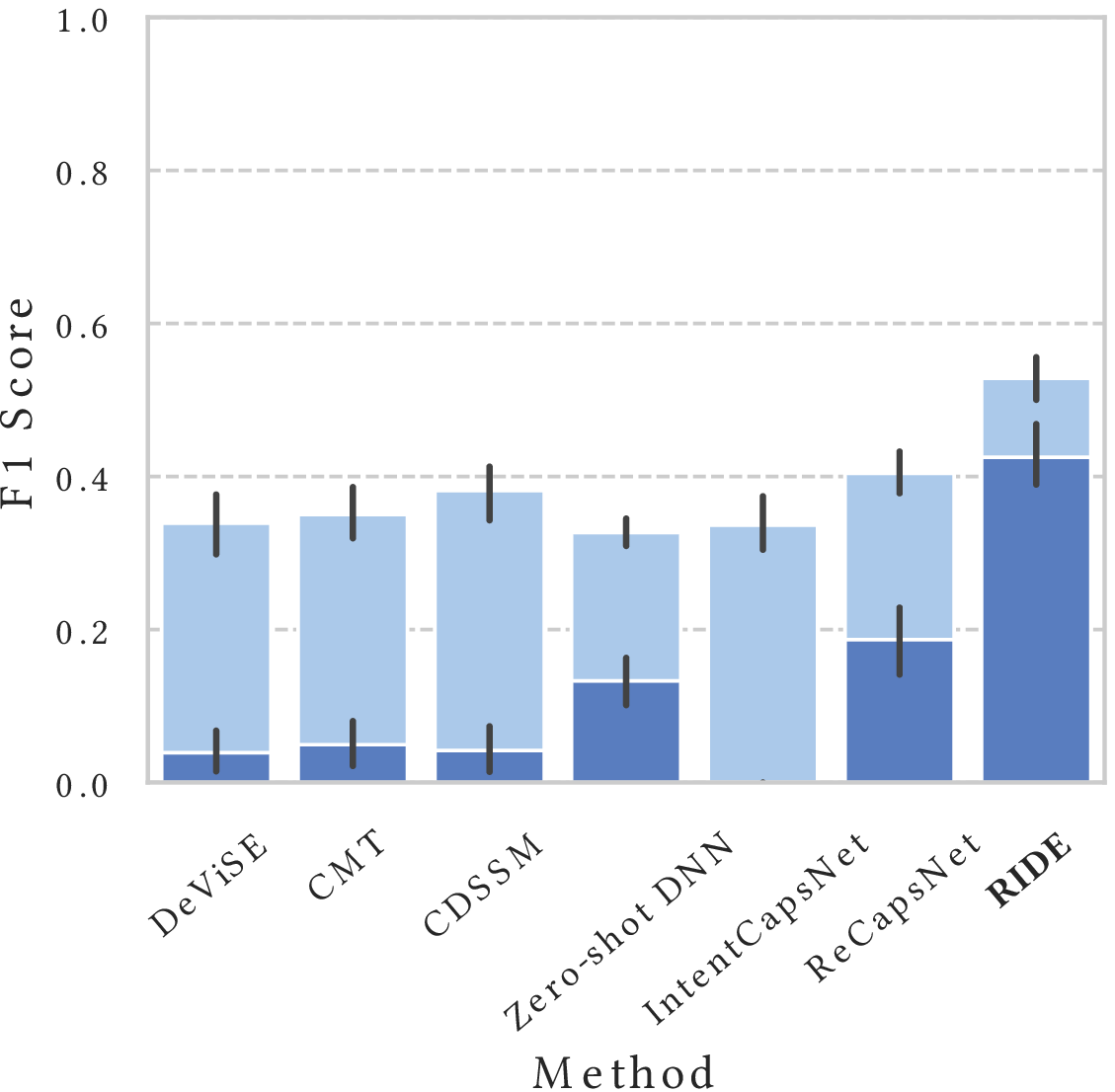}
        \caption{MultiWOZ dataset}
    \end{subfigure} 
    \caption{F1 scores for unseen intents for the competing models in the GZS setting after integrating a PU classifier. 
    }
\label{fig:pu}
%\vspace{-5pt}
\end{figure*}

\section{Results}
\label{results}
\stitle{Standard ZS Intent Detection.}
Figure~\ref{fig:zs} presents the F1 scores averaged over $10$ runs for all competing models with varying percentages of seen intents in the ZS setting.
The performance of all models improves as the percentage of seen intents increases, which is expected because increasing the percentage of seen intent gives models access to more training data and intents.
Our model {\OURMODEL} consistently outperforms the SOTA model SEG~\cite{yan2020unknown} and all other models in the ZS setting with a large margin across all the datasets.
Specifically, it is at least $12.65\%$ more accurate on F1 score than the second best model for any percentage of seen intents on all the datasets.
Note that all models perform worse on SGD and MultiWOZ compared to SNIPS because these two datasets are more challenging: they contain closely related intent labels such as ``FindRestaurant'' and ``FindHotel''.

\stitle{GZS Intent Detection.}
Table~\ref{tab:generalized} shows accuracy and F1 scores averaged over $10$ runs for all competing models in the GZS setting.
For unseen intents, our model {\OURMODEL} outperforms all other competing models on accuracy with a large margin.
Specifically, {\OURMODEL} is $30.36\%$, $58.50\%$, and $53.18\%$ more accurate than the SOTA model SEG on SNIPS, SGD, and MultiWOZ for unseen intents, respectively. Moreover, our model consistently achieves the highest F1 score on seen as well as unseen intents, which confirms its generalizability.
CMT and IntentCapsNet achieve the highest accuracy for utterances with seen intents on all datasets, but their F1 score is among the worst due to their biased-ness towards misclassifying utterances with unseen intents into seen ones.
{\OURMODEL} outperforms the SOTA model SEG regardless of whether a PU classifier is incorporated or not.
For SNIPS, the role of the PU classifier is negligible as it causes a slight improvement in accuracy and F1 score. For SGD and MultiWOZ, which are more challenging datasets, the PU classifier is responsible for significant improvements in accuracy. Specifically, it provides $20.36$ and $19.82$ percentage points improvement for SGD and MultiWOZ, respectively, on unseen intents. 

\stitle{Effect of PU Classifier on Other Models.}
We observed that one of the main sources of error for most models in the GZS setting is their tendency to misclassify utterances with unseen intents into seen ones due to overfitting to seen intents. We investigated whether existing models can be adapted to accurately classify utterances with unseen intents by partially eliminating their bias towards seen intents. Figure~\ref{fig:pu} presents F1 scores of all the models with and without PU classifier. 
A PU classifier significantly improves the results of all the competing models. For instance, the IntentCapsNet model with a PU classifier achieves an F1 score of 74\% for unseen intents on SNIPS dataset in the GZS setting compared to an F1 score of less than 0.01\% without the PU classifier.
Note that the PU classifier has an accuracy (i.e., correctly predicting whether the utterance implies a seen or an unseen intent) of $93.69$ and an F1 score of $93.75$ for SNIPS dataset; $86.13$ accuracy and $83.54$ F1 score for SGD dataset; and $87.32$ accuracy and $88.51$ F1 score for MultiWOZ dataset.
Interestingly, our model {\OURMODEL} (without PU classifier) outperforms all the competing models even when a PU classifier is incorporated into them, which highlights that the PU classifier is not the main source of the performance of our model.
We did not incorporate the PU classifier into SEG model because it already incorporates an equivalent mechanism to distinguish seen intents from unseen ones (i.e., outlier detection).

\begin{table}[t!]
\footnotesize
\centering
\begin{tabular}{l|ccc}
\hline
\textbf{Configuration} & \textbf{SNIPS} & \textbf{SGD} & \textbf{MultiWOZ} \\ \hline
UI-Embed ${\NOPU}$              & 0.2367             & 0.1578             & 0.1723             \\
Rel-M       ${\NOPU}$           & 0.7103             & 0.3593             & 0.3321             \\
RIDE ${\NOPU}$           & \underline{0.9103}             & 0.4634             & 0.4645             \\
\hline
UI-Embed ${\WITHPU}$         & 0.7245             & 0.4202             & 0.4124             \\
Rel-M ${\WITHPU}$            & 0.8463             & \underline{0.5167}             & \underline{0.4781}             \\
RIDE ${\WITHPU}$             &    \textbf{0.9254}          & \textbf{0.5734}             & \textbf{0.5206}             \\ \hline
\end{tabular}
\caption{Ablation study: F1 scores for unseen intents in GZS setting; the key reason behind our model's astonishing accuracy is our relationship meta-features.}
\label{tab:ablation}
%\vspace{-5pt}
\end{table}

\stitle{Ablation Study.}
To quantify the effectiveness of each component in our model, we present the results of our ablation study in Table~\ref{tab:ablation} (in the GZS setting).
Utilizing utterance and intent embeddings only (i.e., UI-Embed) results in very low F1 score, i.e., 23.67\% on SNIPS dataset. Employing relationship meta-features only (i.e., Rel-M) results in significantly better results: an F1 score of 71.03\% on SNIPS dataset. When utterance and intent embeddings are used in conjunction with relationship meta-features (i.e., {\OURMODEL}~${\NOPU}$), it achieves better F1 score compared to the Rel-M or UI-Embed configurations. A similar trend can be observed for the other datasets as well.
Finally, when our entire model is deployed (i.e., including utterance and intent embeddings, relationship meta-features, and the PU classifier, i.e., {\OURMODEL}~${\WITHPU}$), it achieves the best results on all the datasets.

%% file: sections/related.tex
\section{Related Work}
\label{related}
The deep neural networks have proved highly effective for many critical NLP tasks~\cite{pupkdd2020,aarsynthbigdata2020,zhang2018joint,williams2019zero,ma2019end,leonawww21,liu2016attention,gupta2019simple}.
We organize the related work on intent detection into three categories: \myNum{i} supervised intent detection, \myNum{ii} standard zero-shot intent detection, and \myNum{iii} generalized zero-shot intent detection.

\stitle{Supervised Intent Detection.}
%The authors in~\cite{kim2016intent} proposed a model for supervised intent detection using semantic lexicon-enriched word embeddings, and recurrent neural networks were employed in~\cite{ravuri2015recurrent}. 
Recurrent neural networks~\cite{ravuri2015recurrent} and semantic lexicon-enriched word embeddings~\cite{kim2016intent} have been employed for supervised intent detection.
Recently, researchers have proposed solving the related problems of intent detection and slot-filling jointly~\cite{liu2016attention, zhang2018joint, xu2013convolutional}. 
Supervised intent classification works assume the availability of a large amount of labeled training data for all intents to learn discriminative features, whereas we focus on the more challenging and more practically relevant setting where intents are evolving and training data is not available for all intents.

\stitle{Standard Zero-shot Intent Detection.}
The authors in~\cite{yazdani2015model} proposed using label ontologies~\cite{ferreira2015online} (i.e., manually annotated intent attributes) to facilitate generalizing a model to support unseen intents.
The authors in~\cite{dauphin2013zero, kumar2017zero, williams2019zero} map utterances and intents to the same semantic vector space, and then classify utterances based on their proximity to intent labels in that space. Similarly, the authors in~\cite{gangal2019likelihood} employ the outlier detection algorithm LOF~\cite{breunig2000lof} and likelihood ratios for identifying out-of-domain test examples.
While these works showed promising results for intent detection when training data is unavailable for some intents, they assume that all utterances faced at inference time imply unseen intents only. Extending such works to remove the aforementioned assumption is nontrivial. Our model does not assume knowledge of whether an utterance implies a seen or an unseen intent at inference time.

\stitle{Generalized Zero-shot Intent Detection.}
To the best of our knowledge, the authors in~\cite{liu2019reconstructing} proposed the first work that specifically targets the GZS intent detection setting.
They attempt to make their model generalizable to unseen intents by adding a dimensional attention module to a capsule network and learning generalizable transformation matrices from seen intents. 
Recently, the authors in~\cite{yan2020unknown} proposed using a density-based outlier detection algorithm LOF~\cite{breunig2000lof} and semantic-enhanced Gaussian mixture model with large margin loss to learn class-concentrated embeddings to detect unseen intents. In contrast, we leverage rich commonsense knowledge graph to capture deep semantic and discriminative relationships between utterances and intents, which significantly reduces the bias towards classifying unseen intents into seen ones.
In a related, but orthogonal, line of research, the authors in~\cite{ma2019end,li2020sppd,gulyaev2020goal} addressed the problem of intent detection in the context of dialog state tracking where dialog state and conversation history are available in addition to an input utterance. In contrast, this work and the SOTA models we compare against in our experiments only consider an utterance without having access to any dialog state elements.

%% file: sections/conclusion.tex
\section{Conclusion}
\label{conclusion}
We have presented an accurate generalized zero-shot intent detection model.
Our extensive experimental analysis on three intent detection benchmarks shows that our model is 30.36\% to 58.50\% more accurate than the SOTA model for unseen intents.
The main novelty of our model is its utilization of relationship meta-features to accurately identify matching utterance-intent pairs with very limited reliance on training data, and without making any assumption on whether utterances imply seen or unseen intents at inference time.
Furthermore, our idea of integrating Positive-Unlabeled learning in GZS intent detection models further improves our models' performance, and significantly improves the accuracy of existing models as well.

%% file: sections/appendix.tex
\section{Appendices}
\label{sec:appendix}
This section provides supplementary details on various aspects of this paper. First, we provide more details on the commonsense knowledge graph we have used as a source of knowledge on semantic relatedness of concepts. Then, we describe the specifics of the datasets we used in our evaluation and our preprocessing procedures.
Finally, we provide the details on handling a special case when utterances do not imply intents.

\subsection{Knowledge Graph Details}
\label{sec:kg_details}
Although creating and maintaining knowledge graphs is laborious and time consuming, the immense utility of such graphs has led many researchers and institutions to make the effort of building and maintaining knowledge graphs in many domains, which lifts the burden off of other researchers and developers who utilize these graphs. For tasks that involve commonsense reasoning such as \emph{generalized zero-shot} intent detection, the ConceptNet~\cite{speer2016conceptnet} commonsense knowledge graph stands out as one of the most popular and freely available resources.
ConceptNet originated from the crowdsourcing project Open Mind Common Sense, and includes knowledge not only from crowdsourced resources but also expert-curated resources. It is available in $10$ core languages, and $68$ more common languages. It was employed to show state-of-the-art results at SemEval 2017~\cite{speer2017conceptnet}.
In this work, we considered $35$ relation types to generate our relationship meta-features. The relation types are:
\GEDGE{RelatedTo}, \GEDGE{FormOf}, \GEDGE{IsA}, \GEDGE{PartOf}, \GEDGE{HasA}, \GEDGE{UsedFor}, \GEDGE{CapableOf}, \GEDGE{AtLocation}, \GEDGE{Causes}, \GEDGE{HasSubevent}, \GEDGE{HasFirstSubevent}, \GEDGE{HasLastSubevent}, \GEDGE{HasPrerequisite}, \GEDGE{HasProperty}, \GEDGE{MotivatedByGoal}, \GEDGE{ObstructedBy}, \GEDGE{Desires}, \GEDGE{CreatedBy}, \GEDGE{Synonym}, \GEDGE{Antonym}, \GEDGE{DistinctFrom}, \GEDGE{DerivedFrom}, \GEDGE{SymbolOf}, \GEDGE{DefinedAs}, \GEDGE{MannerOf}, \GEDGE{LocatedNear}, \GEDGE{HasContext}, \GEDGE{SimilarTo}, \GEDGE{EtymologicallyRelatedTo}, \GEDGE{EtymologicallyDerivedFrom}, \GEDGE{CausesDesire}, \GEDGE{MadeOf}, \GEDGE{ReceivesAction}, \GEDGE{ExternalURL}, and \GEDGE{Self}. 

The relationship meta-feature generator produces $35 \times 4 = 140$ dimension vector for each utterance-intent pair. Specifically, we generate relationships:
\myNum{i} from utterance to Object (i.e., Object part in intent label);
\myNum{ii} utterance to Action (i.e., Action part in intent label); and
\myNum{iii} Object to utterance;
\myNum{iv} Action to utterance.

A knowledge graph may not have redundant, but necessary information. For example, a knowledge graph may have the entry \KGT{movie}{IsA}{film} but not \KGT{film}{IsA}{movie} or vice-versa, because one triple can be inferred from the other based on background knowledge (i.e., symmetric nature of the \GEDGE{IsA} relation). Similarly, the triple \KGT{movie}{HasA}{subtitles} can be used to infer the triple \KGT{subtitles}{PartOf}{movie} based on the background knowledge (i.e., inverse relation between \GEDGE{HasA} and \GEDGE{PartOf}).
So, if this kind of redundant information (i.e., complementing entries for all such triples) is not available in the knowledge graph itself, there is no way for the model to learn these relationships automatically.
To overcome this issue, we incorporate the background knowledge that each of the relation types 
\GEDGE{IsA}, 
\GEDGE{RelatedTo},
\GEDGE{Synonym},
\GEDGE{Antonym},
\GEDGE{DistinctFrom},
\GEDGE{LocatedNear},
\GEDGE{SimilarTo}, and
\GEDGE{EtymologicallyRelatedTo} is symmetric; and that
the relation types \GEDGE{PartOf} and \GEDGE{HasA} are inversely related in our link prediction model as described in~\cite{kazemi2018simple}.

\subsection{Training the Link Predictor}
\label{sec:app:lp}
The training data for a link prediction model is prepared as follows. First, the triples in the input knowledge graph are assigned a label of $1$. Then, negative examples are generated by corrupting true triples (i.e., modifying the \GNODE{head} or \GNODE{tail} of existing triples) and assigning them a label of $-1$~\cite{bordes2013translating}.
Finally, we train our LP using the generated training data by minimizing the L$2$ regularized negative log-likelihood loss of training triples~\cite{trouillon2016complex}.

\subsection{Datasets Preprocessing}
\label{sec:preprocessing}

SNIPS Natural Language Understanding benchmark (SNIPS)~\cite{coucke2018snips} is a commonly used dataset for intent detection, whereas Dialogue System Technology Challenge 8, Schema Guided Dialogue dataset (SGD)~\cite{rastogi2019towards} and Multi-Domain Wizard-of-Oz (MultiWOZ)~\cite{zang-etal-2020-multiwoz} were originally proposed for the task of dialogue state tracking. For SGD and MultiWOZ, we perform a few trivial preprocessing steps to extract utterances that contain intents, along with their labels, and use them for the task of generalized zero-shot intent detection.
First, we provide the details on the preprocessing steps specific to the SGD and MultiWOZ dataset and then describe the preprocessing steps that are common for all datasets.

\stitle{Steps for SGD and MultiWOZ.}
To maintain consistency with the previous work on intent detection, we extract only the utterances where user/system expresses an intent, and discard the rest from the original SGD and MultiWOZ datasets. The dialogue state contains a property ``active\_intent'' that keeps track of the user's current intent. After each user utterance, we compare dialogue states to check for the expression of a new user intent, i.e., whether the value of the ``active\_intent'' is modified.  
Whenever the user expresses a new intent, the value of the ``active\_intent'' is updated. 
Moreover, sometimes, the bot (i.e., system) also offers new intents to the user (e.g., offering reserving a table to the user, who has successfully searched for a restaurant), which is tracked in the system actions property ``act =  OFFER\_INTENT'', and ``values = $<$new\_intent$>$''. We also keep such system utterances. 

\stitle{Common Steps.}
We perform some standard preprocessing steps on all the datasets. We use spaCy to tokenize the sentences.
Since intent labels are given in the ``ActionObject'' format, we tokenize them into ``Action Object'' phrases before feeding them into our model.
For example, the intent labels ``FindHotel'' and ``RateBook'', are transformed into ``Find Hotel'' and ``Rate Book'', respectively.
Note that some Objects parts of intent labels are compound. Consider the intent label ``SearchOneWayFlight'' whose Action is ``Search'' and Object is ``OneWayFlight''. In such cases, our relationship meta-features generator computes meta-features for each part of the compound object then averages them to produce the Object meta-features vector. In the previous example, ``OneWayFlight'' meta-features vector is computed as the average of the meta-features of ``OneWay'' and ``Flight''.

\subsection{Intent Existence Prediction}
In real human-to-human or human-to-machine conversations, utterances do not necessarily imply intents. Most existing intent detection models formulate the problem as a classification problem where utterances are assumed to imply an intent, which limits utility of such models in practice. In what follows, we describe a simple method for extending intent detection models (including ours) to accommodate the case when utterances do not necessarily imply intents.
We propose to do binary classification as a first step in intent detection, where a binary classifier is used to identify utterances that do not imply an intent. To validate the viability of this proposal, we experimented with several binary classifiers. To train the classifiers, we created datasets of positive and negative examples from seen intents data; positive examples are utterances that imply intents, and negative examples are utterances that do not have intents (See Section~\ref{sec:preprocessing} for details on identifying utterances that imply intents). For the SGD dataset, we used the standard train/test splits, and for the MultiWOZ dataset, we used the same splits described in the GZS setting.
We report in Table~\ref{tab:intent-existence} the average F1 score over 5 runs of several binary classifiers for the SGD and the MultiWOZ datasets. All classifiers use ELMo~\cite{Peters:2018} and POS tag embeddings. These results show that intent existence classification can be done accurately using the available training data; consequently, intent detection models can be easily and reliably extended to support the case when some input utterances do not imply intents.

\begin{table}[t!]
\centering
\footnotesize
\begin{tabular}{l|cc}
\hline
\textbf{Method} & \textbf{SGD} & \textbf{MultiWOZ} \\ \hline
CNN             & 0.9497       & 0.9512             \\
GRU             & \textbf{0.9528}       & \underline{0.9619}             \\
LSTM            & 0.9512       & 0.9607              \\
Bi-LSTM         & \underline{0.9525}       & \textbf{0.9621}  \\
\hline
\end{tabular}
\caption{F1 score for intent existence binary classifiers.}
\label{tab:intent-existence}
\end{table}

%% file: main.bbl
\begin{thebibliography}{42}
\expandafter\ifx\csname natexlab\endcsname\relax\def\natexlab#1{#1}\fi

\bibitem[{Bordes et~al.(2013)Bordes, Usunier, Garcia-Duran, Weston, and
  Yakhnenko}]{bordes2013translating}
Antoine Bordes, Nicolas Usunier, Alberto Garcia-Duran, Jason Weston, and Oksana
  Yakhnenko. 2013.
\newblock Translating embeddings for modeling multi-relational data.
\newblock In \emph{Advances in neural information processing systems}, pages
  2787--2795.

\bibitem[{Breunig et~al.(2000)Breunig, Kriegel, Ng, and
  Sander}]{breunig2000lof}
Markus~M Breunig, Hans-Peter Kriegel, Raymond~T Ng, and J{\"o}rg Sander. 2000.
\newblock Lof: identifying density-based local outliers.
\newblock In \emph{Proceedings of the 2000 ACM SIGMOD international conference
  on Management of data}, pages 93--104.

\bibitem[{Cer et~al.(2018)Cer, Yang, Kong, Hua, Limtiaco, John, Constant,
  Guajardo-Cespedes, Yuan, Tar et~al.}]{cer2018universal}
Daniel Cer, Yinfei Yang, Sheng-yi Kong, Nan Hua, Nicole Limtiaco, Rhomni~St
  John, Noah Constant, Mario Guajardo-Cespedes, Steve Yuan, Chris Tar, et~al.
  2018.
\newblock Universal sentence encoder.
\newblock \emph{arXiv preprint arXiv:1803.11175}.

\bibitem[{Chen et~al.(2016)Chen, Hakkani-T{\"u}r, and He}]{chen2016zero}
Yun-Nung Chen, Dilek Hakkani-T{\"u}r, and Xiaodong He. 2016.
\newblock Zero-shot learning of intent embeddings for expansion by
  convolutional deep structured semantic models.
\newblock In \emph{2016 IEEE International Conference on Acoustics, Speech and
  Signal Processing (ICASSP)}, pages 6045--6049. IEEE.

\bibitem[{Chen et~al.(2013)Chen, Liu, Hsu, Castellanos, and
  Ghosh}]{chen2013identifying}
Zhiyuan Chen, Bing Liu, Meichun Hsu, Malu Castellanos, and Riddhiman Ghosh.
  2013.
\newblock Identifying intention posts in discussion forums.
\newblock In \emph{Proceedings of the 2013 conference of the North American
  chapter of the association for computational linguistics: human language
  technologies}, pages 1041--1050.

\bibitem[{Coucke et~al.(2018)Coucke, Saade, Ball, Bluche, Caulier, Leroy,
  Doumouro, Gisselbrecht, Caltagirone, Lavril et~al.}]{coucke2018snips}
Alice Coucke, Alaa Saade, Adrien Ball, Th{\'e}odore Bluche, Alexandre Caulier,
  David Leroy, Cl{\'e}ment Doumouro, Thibault Gisselbrecht, Francesco
  Caltagirone, Thibaut Lavril, et~al. 2018.
\newblock Snips voice platform: an embedded spoken language understanding
  system for private-by-design voice interfaces.
\newblock \emph{arXiv preprint arXiv:1805.10190}.

\bibitem[{Dauphin et~al.(2013)Dauphin, Tur, Hakkani-Tur, and
  Heck}]{dauphin2013zero}
Yann~N Dauphin, Gokhan Tur, Dilek Hakkani-Tur, and Larry Heck. 2013.
\newblock Zero-shot learning for semantic utterance classification.
\newblock \emph{arXiv preprint arXiv:1401.0509}.

\bibitem[{Elkan and Noto(2008)}]{elkan2008learning}
Charles Elkan and Keith Noto. 2008.
\newblock Learning classifiers from only positive and unlabeled data.
\newblock In \emph{Proceedings of the 14th ACM SIGKDD international conference
  on Knowledge discovery and data mining}, pages 213--220.

\bibitem[{Farooq et~al.(2020)Farooq, Siddique, Jamour, Zhao, and
  Hristidis}]{aarsynthbigdata2020}
Umar Farooq, A.~B. Siddique, Fuad Jamour, Zhijia Zhao, and Vagelis Hristidis.
  2020.
\newblock App-aware response synthesis for user reviews.
\newblock In \emph{2020 IEEE International Conference on Big Data (Big Data)},
  pages 699--708.

\bibitem[{Felix et~al.(2018)Felix, Kumar, Reid, and Carneiro}]{felix2018multi}
Rafael Felix, Vijay~BG Kumar, Ian Reid, and Gustavo Carneiro. 2018.
\newblock Multi-modal cycle-consistent generalized zero-shot learning.
\newblock In \emph{Proceedings of the European Conference on Computer Vision
  (ECCV)}, pages 21--37.

\bibitem[{Ferreira et~al.(2015)Ferreira, Jabaian, and
  Lefevre}]{ferreira2015online}
Emmanuel Ferreira, Bassam Jabaian, and Fabrice Lefevre. 2015.
\newblock Online adaptative zero-shot learning spoken language understanding
  using word-embedding.
\newblock In \emph{2015 IEEE International Conference on Acoustics, Speech and
  Signal Processing (ICASSP)}, pages 5321--5325. IEEE.

\bibitem[{Frome et~al.(2013)Frome, Corrado, Shlens, Bengio, Dean, Ranzato, and
  Mikolov}]{frome2013devise}
Andrea Frome, Greg~S Corrado, Jon Shlens, Samy Bengio, Jeff Dean, Marc'Aurelio
  Ranzato, and Tomas Mikolov. 2013.
\newblock Devise: A deep visual-semantic embedding model.
\newblock In \emph{Advances in neural information processing systems}, pages
  2121--2129.

\bibitem[{Gangal et~al.(2019)Gangal, Arora, Einolghozati, and
  Gupta}]{gangal2019likelihood}
Varun Gangal, Abhinav Arora, Arash Einolghozati, and Sonal Gupta. 2019.
\newblock Likelihood ratios and generative classifiers for unsupervised
  out-of-domain detection in task oriented dialog.
\newblock \emph{arXiv preprint arXiv:1912.12800}.

\bibitem[{Gulyaev et~al.(2020)Gulyaev, Elistratova, Konovalov, Kuratov,
  Pugachev, and Burtsev}]{gulyaev2020goal}
Pavel Gulyaev, Eugenia Elistratova, Vasily Konovalov, Yuri Kuratov, Leonid
  Pugachev, and Mikhail Burtsev. 2020.
\newblock Goal-oriented multi-task bert-based dialogue state tracker.
\newblock \emph{arXiv preprint arXiv:2002.02450}.

\bibitem[{Gupta et~al.(2019)Gupta, Hewitt, and Kirchhoff}]{gupta2019simple}
Arshit Gupta, John Hewitt, and Katrin Kirchhoff. 2019.
\newblock Simple, fast, accurate intent classification and slot labeling for
  goal-oriented dialogue systems.
\newblock In \emph{Proceedings of the 20th Annual SIGdial Meeting on Discourse
  and Dialogue}, pages 46--55.

\bibitem[{Kazemi and Poole(2018)}]{kazemi2018simple}
Seyed~Mehran Kazemi and David Poole. 2018.
\newblock Simple embedding for link prediction in knowledge graphs.
\newblock In \emph{Advances in neural information processing systems}, pages
  4284--4295.

\bibitem[{Kim et~al.(2016)Kim, Tur, Celikyilmaz, Cao, and Wang}]{kim2016intent}
Joo-Kyung Kim, Gokhan Tur, Asli Celikyilmaz, Bin Cao, and Ye-Yi Wang. 2016.
\newblock Intent detection using semantically enriched word embeddings.
\newblock In \emph{2016 IEEE Spoken Language Technology Workshop (SLT)}, pages
  414--419. IEEE.

\bibitem[{Kumar et~al.(2017)Kumar, Muddireddy, Dreyer, and
  Hoffmeister}]{kumar2017zero}
Anjishnu Kumar, Pavankumar~Reddy Muddireddy, Markus Dreyer, and Bj{\"o}rn
  Hoffmeister. 2017.
\newblock Zero-shot learning across heterogeneous overlapping domains.
\newblock In \emph{INTERSPEECH}, pages 2914--2918.

\bibitem[{Li et~al.(2020)Li, Xiong, and Cao}]{li2020sppd}
Miao Li, Haoqi Xiong, and Yunbo Cao. 2020.
\newblock The sppd system for schema guided dialogue state tracking challenge.
\newblock \emph{arXiv preprint arXiv:2006.09035}.

\bibitem[{Liu and Lane(2016)}]{liu2016attention}
Bing Liu and Ian Lane. 2016.
\newblock Attention-based recurrent neural network models for joint intent
  detection and slot filling.
\newblock \emph{arXiv preprint arXiv:1609.01454}.

\bibitem[{Liu et~al.(2019)Liu, Zhang, Fan, Fu, Li, Wu, and
  Lam}]{liu2019reconstructing}
Han Liu, Xiaotong Zhang, Lu~Fan, Xuandi Fu, Qimai Li, Xiao-Ming Wu, and
  Albert~YS Lam. 2019.
\newblock Reconstructing capsule networks for zero-shot intent classification.
\newblock In \emph{Proceedings of the 2019 Conference on Empirical Methods in
  Natural Language Processing and the 9th International Joint Conference on
  Natural Language Processing (EMNLP-IJCNLP)}, pages 4801--4811.

\bibitem[{Ma et~al.(2019)Ma, Zeng, Zhu, Li, Yang, Yao, Zhou, and
  Shen}]{ma2019end}
Yue Ma, Zengfeng Zeng, Dawei Zhu, Xuan Li, Yiying Yang, Xiaoyuan Yao, Kaijie
  Zhou, and Jianping Shen. 2019.
\newblock An end-to-end dialogue state tracking system with machine reading
  comprehension and wide \& deep classification.
\newblock \emph{arXiv preprint arXiv:1912.09297}.

\bibitem[{Peters et~al.(2018)Peters, Neumann, Iyyer, Gardner, Clark, Lee, and
  Zettlemoyer}]{Peters:2018}
Matthew~E. Peters, Mark Neumann, Mohit Iyyer, Matt Gardner, Christopher Clark,
  Kenton Lee, and Luke Zettlemoyer. 2018.
\newblock Deep contextualized word representations.
\newblock In \emph{Proc. of NAACL}.

\bibitem[{PyTorch(2020)}]{torchopt68:online}
PyTorch. 2020.
\newblock torch.optim — pytorch 1.3.0 documentation.
\newblock
  \url{https://pytorch.org/docs/stable/optim.html#torch.optim.lr_scheduler.ReduceLROnPlateau}.
\newblock (Accessed on 11/12/2020).

\bibitem[{Rastogi et~al.(2019)Rastogi, Zang, Sunkara, Gupta, and
  Khaitan}]{rastogi2019towards}
Abhinav Rastogi, Xiaoxue Zang, Srinivas Sunkara, Raghav Gupta, and Pranav
  Khaitan. 2019.
\newblock Towards scalable multi-domain conversational agents: The
  schema-guided dialogue dataset.
\newblock \emph{arXiv preprint arXiv:1909.05855}.

\bibitem[{Ravuri and Stolcke(2015)}]{ravuri2015recurrent}
Suman Ravuri and Andreas Stolcke. 2015.
\newblock Recurrent neural network and lstm models for lexical utterance
  classification.
\newblock In \emph{Sixteenth Annual Conference of the International Speech
  Communication Association}.

\bibitem[{Sabour et~al.(2017)Sabour, Frosst, and Hinton}]{sabour2017dynamic}
Sara Sabour, Nicholas Frosst, and Geoffrey~E Hinton. 2017.
\newblock Dynamic routing between capsules.
\newblock In \emph{Advances in neural information processing systems}, pages
  3856--3866.

\bibitem[{Siddique et~al.(2020)Siddique, Oymak, and Hristidis}]{pupkdd2020}
A.~B. Siddique, Samet Oymak, and Vagelis Hristidis. 2020.
\newblock \href {https://doi.org/10.1145/3394486.3403231} {Unsupervised
  paraphrasing via deep reinforcement learning}.
\newblock In \emph{Proceedings of the 26th ACM SIGKDD International Conference
  on Knowledge Discovery \& Data Mining}, KDD '20, page 1800–1809, New York,
  NY, USA. Association for Computing Machinery.

\bibitem[{Siddique et~al.(2021)Siddique, Jamour, and Hristidis}]{leonawww21}
A.B. Siddique, Fuad Jamour, and Vagelis Hristidis. 2021.
\newblock \href {https://doi.org/10.1145/3442381.3449870}
  {Linguistically-enriched and context-aware zero-shot slot filling}.
\newblock In \emph{Proceedings of the Web Conference 2021}, New York, NY, USA.
  Association for Computing Machinery.

\bibitem[{Socher et~al.(2013)Socher, Ganjoo, Manning, and Ng}]{socher2013zero}
Richard Socher, Milind Ganjoo, Christopher~D Manning, and Andrew Ng. 2013.
\newblock Zero-shot learning through cross-modal transfer.
\newblock In \emph{Advances in neural information processing systems}, pages
  935--943.

\bibitem[{Speer et~al.(2016)Speer, Chin, and Havasi}]{speer2016conceptnet}
Robyn Speer, Joshua Chin, and Catherine Havasi. 2016.
\newblock Conceptnet 5.5: An open multilingual graph of general knowledge.
\newblock \emph{arXiv preprint arXiv:1612.03975}.

\bibitem[{Speer et~al.(2017)Speer, Chin, and Havasi}]{speer2017conceptnet}
Robyn Speer, Joshua Chin, and Catherine Havasi. 2017.
\newblock \href {http://aaai.org/ocs/index.php/AAAI/AAAI17/paper/view/14972}
  {Conceptnet 5.5: An open multilingual graph of general knowledge}.

\bibitem[{Trouillon et~al.(2016)Trouillon, Welbl, Riedel, Gaussier, and
  Bouchard}]{trouillon2016complex}
Th{\'e}o Trouillon, Johannes Welbl, Sebastian Riedel, {\'E}ric Gaussier, and
  Guillaume Bouchard. 2016.
\newblock Complex embeddings for simple link prediction.
\newblock International Conference on Machine Learning (ICML).

\bibitem[{Vedula et~al.(2020)Vedula, Lipka, Maneriker, and
  Parthasarathy}]{vedula2020open}
Nikhita Vedula, Nedim Lipka, Pranav Maneriker, and Srinivasan Parthasarathy.
  2020.
\newblock Open intent extraction from natural language interactions.
\newblock In \emph{Proceedings of The Web Conference 2020}, pages 2009--2020.

\bibitem[{Wang et~al.(2015)Wang, Cong, Zhao, and Li}]{aaaiactobj}
Jinpeng Wang, Gao Cong, Wayne~Xin Zhao, and Xiaoming Li. 2015.
\newblock Mining user intents in twitter: A semi-supervised approach to
  inferring intent categories for tweets.
\newblock In \emph{Proceedings of the Twenty-Ninth AAAI Conference on
  Artificial Intelligence}, AAAI’15, page 318–324. AAAI Press.

\bibitem[{Williams(2019)}]{williams2019zero}
Kyle Williams. 2019.
\newblock Zero shot intent classification using long-short term memory
  networks.
\newblock \emph{Proc. Interspeech 2019}, pages 844--848.

\bibitem[{Xia et~al.(2018)Xia, Zhang, Yan, Chang, and Yu}]{xia2018zero}
Congying Xia, Chenwei Zhang, Xiaohui Yan, Yi~Chang, and Philip~S Yu. 2018.
\newblock Zero-shot user intent detection via capsule neural networks.
\newblock \emph{arXiv preprint arXiv:1809.00385}.

\bibitem[{Xu and Sarikaya(2013)}]{xu2013convolutional}
Puyang Xu and Ruhi Sarikaya. 2013.
\newblock Convolutional neural network based triangular crf for joint intent
  detection and slot filling.
\newblock In \emph{2013 ieee workshop on automatic speech recognition and
  understanding}, pages 78--83. IEEE.

\bibitem[{Yan et~al.(2020)Yan, Fan, Li, Liu, Zhang, Wu, and
  Lam}]{yan2020unknown}
Guangfeng Yan, Lu~Fan, Qimai Li, Han Liu, Xiaotong Zhang, Xiao-Ming Wu, and
  Albert~YS Lam. 2020.
\newblock Unknown intent detection using gaussian mixture model with an
  application to zero-shot intent classification.
\newblock In \emph{Proceedings of the 58th Annual Meeting of the Association
  for Computational Linguistics}, pages 1050--1060.

\bibitem[{Yazdani and Henderson(2015)}]{yazdani2015model}
Majid Yazdani and James Henderson. 2015.
\newblock A model of zero-shot learning of spoken language understanding.
\newblock In \emph{Proceedings of the 2015 Conference on Empirical Methods in
  Natural Language Processing}, pages 244--249.

\bibitem[{Zang et~al.(2020)Zang, Rastogi, Sunkara, Gupta, Zhang, and
  Chen}]{zang-etal-2020-multiwoz}
Xiaoxue Zang, Abhinav Rastogi, Srinivas Sunkara, Raghav Gupta, Jianguo Zhang,
  and Jindong Chen. 2020.
\newblock \href {https://doi.org/10.18653/v1/2020.nlp4convai-1.13}
  {{M}ulti{WOZ} 2.2 : A dialogue dataset with additional annotation corrections
  and state tracking baselines}.
\newblock In \emph{Proceedings of the 2nd Workshop on Natural Language
  Processing for Conversational AI}, pages 109--117, Online. Association for
  Computational Linguistics.

\bibitem[{Zhang et~al.(2018)Zhang, Li, Du, Fan, and Yu}]{zhang2018joint}
Chenwei Zhang, Yaliang Li, Nan Du, Wei Fan, and Philip~S Yu. 2018.
\newblock Joint slot filling and intent detection via capsule neural networks.
\newblock \emph{arXiv preprint arXiv:1812.09471}.

\end{thebibliography}
